\documentclass[final]{cvpr}

\usepackage{times}
\usepackage{epsfig}
\usepackage{graphicx}
\usepackage{amsmath}
\usepackage{amssymb}
\usepackage{subfigure}
\usepackage{boldline}
\usepackage{booktabs}
\usepackage[x11names, table]{xcolor}

\usepackage[pagebackref=true,breaklinks=true,colorlinks,bookmarks=false]{hyperref}

\def\cvprPaperID{****} % *** Enter the CVPR Paper ID here
\def\confYear{CVPR 2021}

\begin{document}

%%%%%%%%% TITLE
\title{Multi Receptive Field Network for Semantic Segmentation}

\author{
Jianlong Yuan \quad Zelu Deng \quad Shu Wang \quad Zhenbo Luo\\
Samsung Research China Beijing\\
{\tt\small yjl9122@163.com, zelu.deng@stu.uestc.edu.cn } \\ 
{\tt\small s0913.wang@samsung.com, luozhenbo@tsinghua.org.cn}
}

\maketitle

%%%%%%%%% ABSTRACT
\begin{abstract}
Semantic segmentation is one of the key tasks in computer vision, which is to assign a category label to each pixel in an image. Despite significant progress achieved recently, most existing methods still suffer from two challenging issues: 1) the size of objects and stuff in an image can be very diverse, demanding for incorporating multi-scale features into the fully convolutional networks (FCNs); 2) the pixels close to or at the boundaries of object/stuff are hard to classify due to the intrinsic weakness of convolutional networks. To address the first issue, we propose a new Multi-Receptive Field Module (MRFM), explicitly taking multi-scale features into account. For the second issue, we design an edge-aware loss which is effective in distinguishing the boundaries of object/stuff. With these two designs,  our Multi Receptive Field Network achieves new state-of-the-art results on two widely-used semantic segmentation benchmark datasets. Specifically, we achieve a mean IoU of 83.0\% on the Cityscapes dataset and 88.4\% mean IoU on the Pascal VOC2012 dataset.
\end{abstract}

%%%%%%%%% BODY TEXT
\section{Introduction}

The task of semantic segmentation is one of the key technologies in visual understanding,
which is widely used in object parsing, scene parsing, human body parsing and automatic driving, etc. The task is to predict each pixel in the image into a prescribed set of categories \cite{caesar2016coco, cityscpaes, coco,everingham2012pascal,SBD}, which is a dense per-pixel prediction task. In recent years, compared with systems relying on hand-crafted features \cite{he2004multiscale,shotton2009textonboost,kohli2009robust,russell2009associative,gould2009decomposing,yao2012describing}, semantic segmentation methods based on deep FCN \cite{FCN1} have made tremendous progress, and these methods have achieved very impressive results on semantic segmentation benchmarks. Despite the success, FCNs still have two limitations.

\begin{figure}[t]
\setlength{\abovecaptionskip}{0.2cm}
\setlength{\belowcaptionskip}{-0.4cm}
\centering
\includegraphics[width=0.98\linewidth]{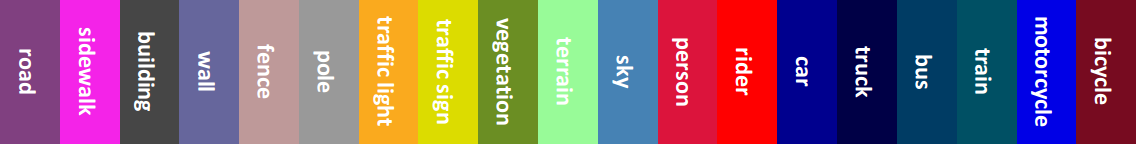} \\
\vspace{-0.15cm}

\subfigure[Image]{
\hspace{-0.15cm}
\begin{minipage}{0.47\linewidth}
\includegraphics[width=\linewidth]{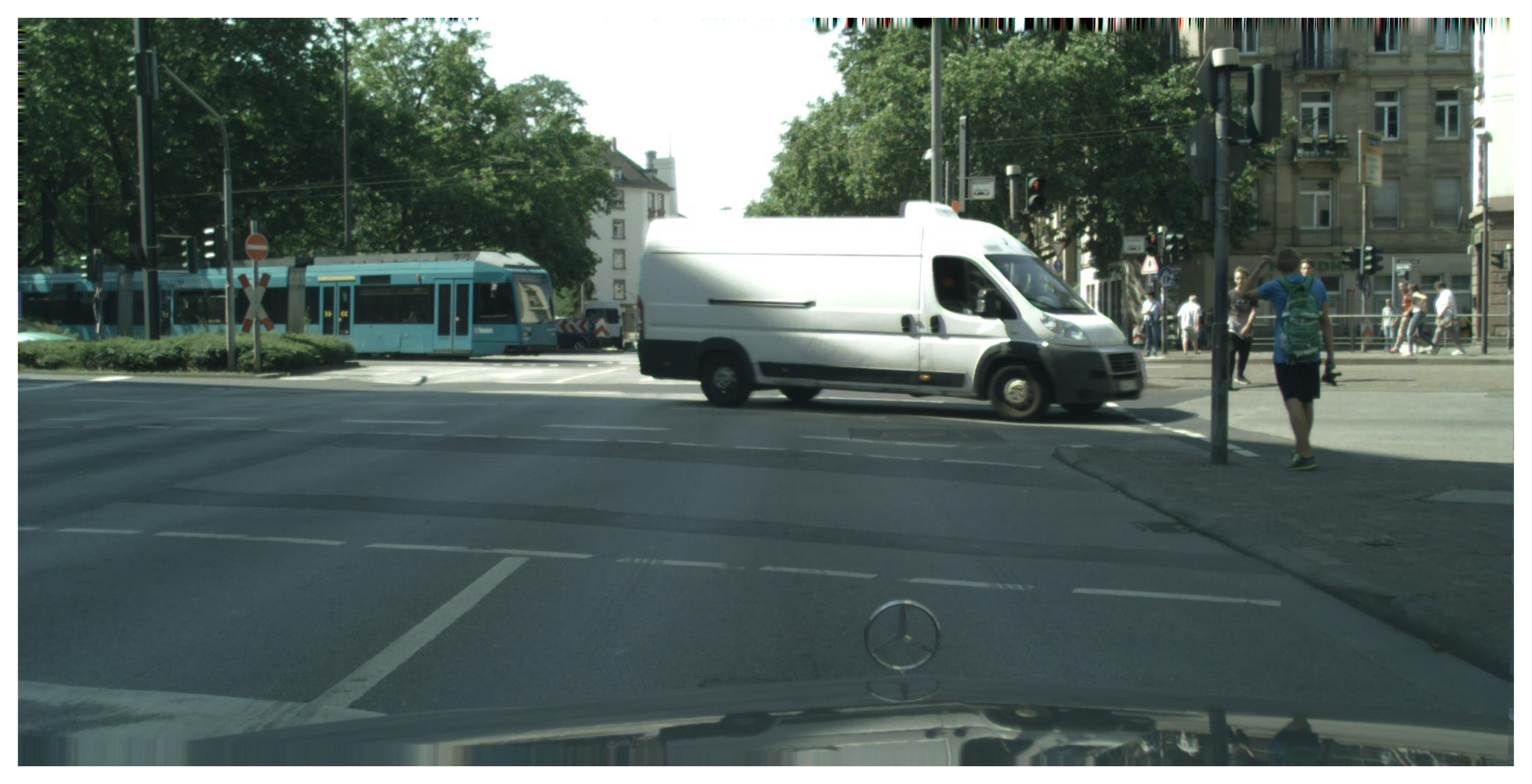}
\includegraphics[width=\linewidth]{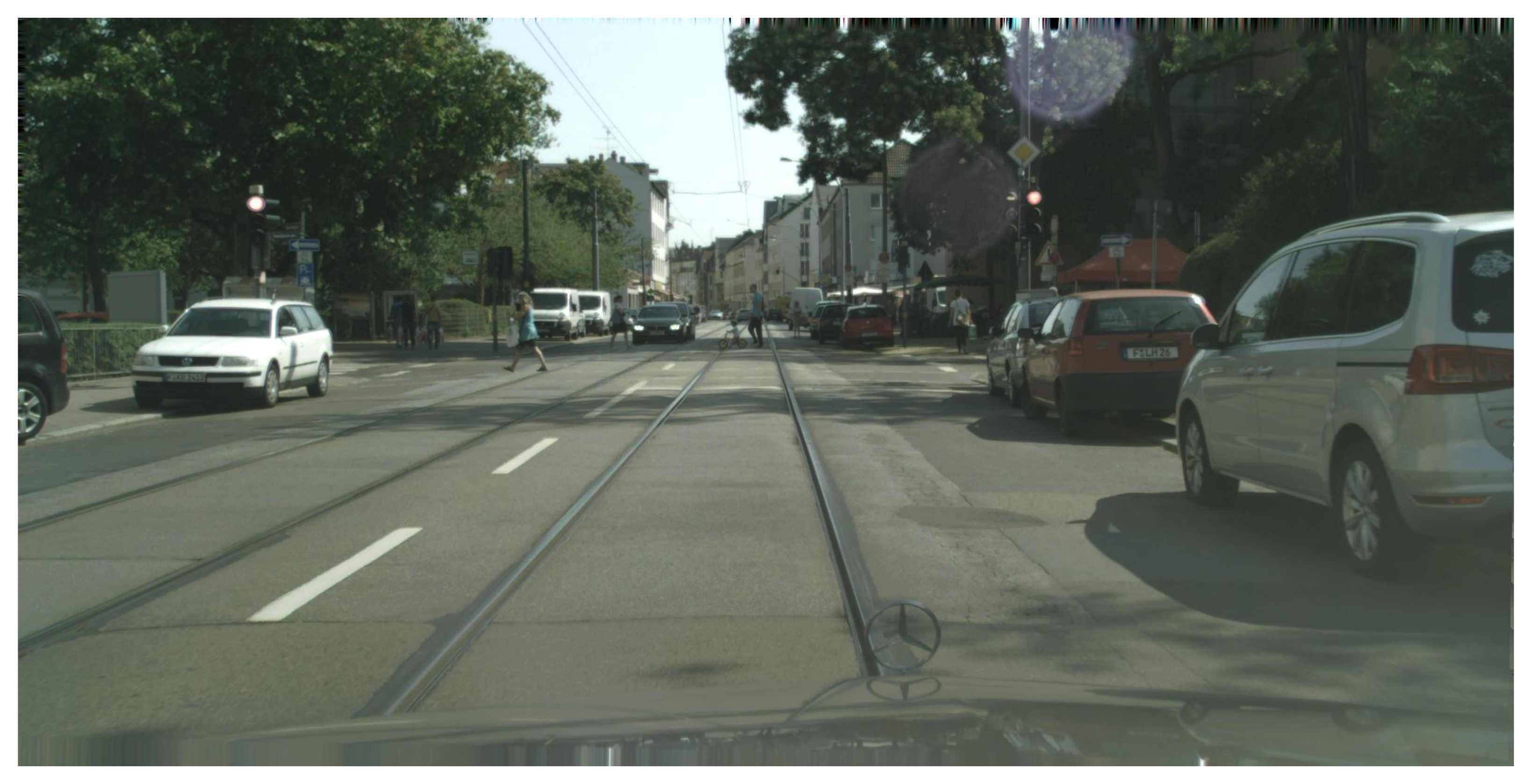}
\end{minipage}
\label{introduction_img}
}
\hspace{0.01cm}
\subfigure[Ground Truth]{
\hspace{-0.15cm}
\begin{minipage}{0.47\linewidth}
\includegraphics[width=\linewidth]{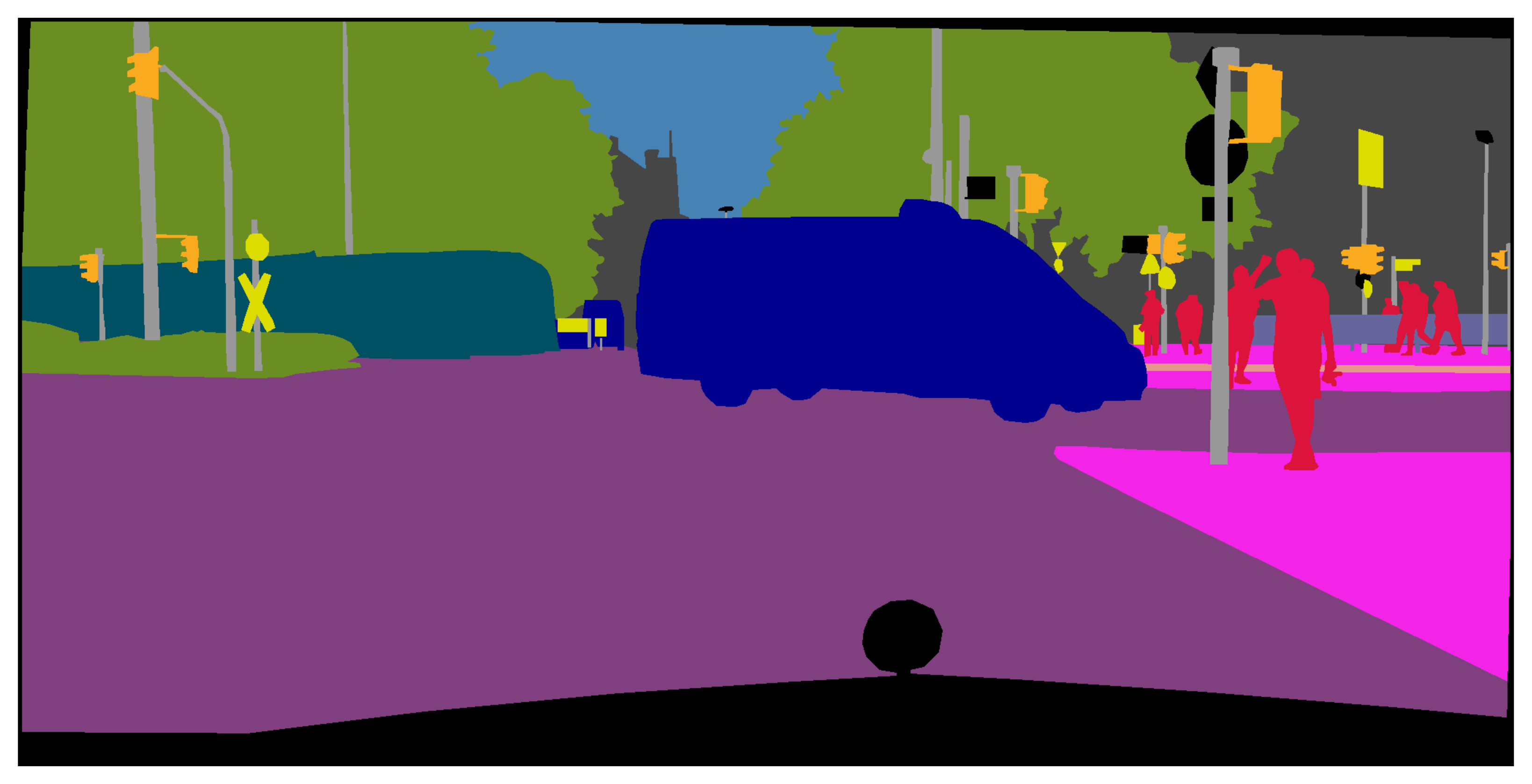}
\includegraphics[width=\linewidth]{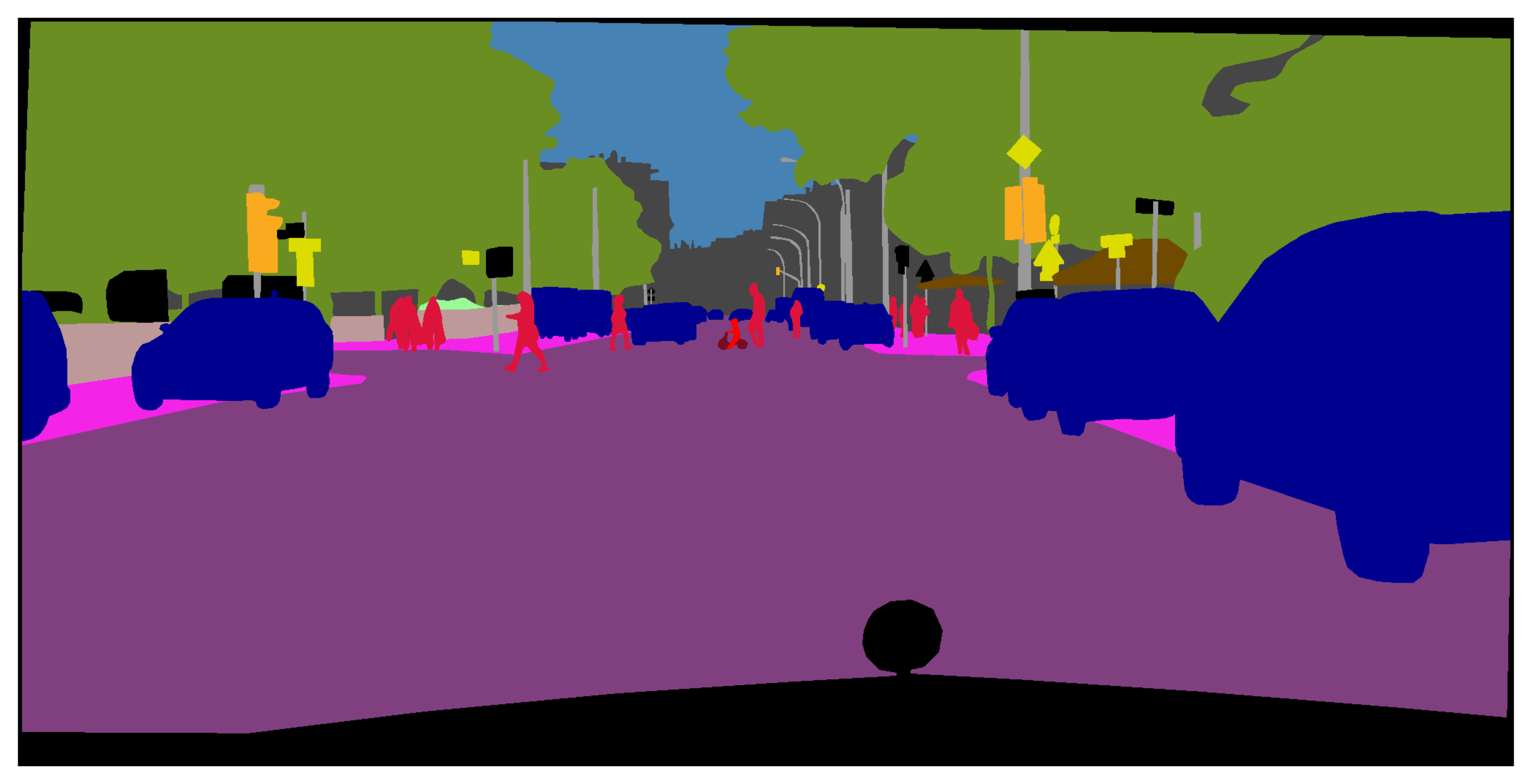}

\end{minipage}
\label{introduction_gt}
}

\caption{Illustration of challenging scenes in the Cityscpaes dataset \cite{cityscpaes}.
Various scales and indistinguishable boundaries of objects/stuff makes it challenging to accurately
parse each pixel.}
\end{figure}

First, the sizes of objects/stuff in images can be very diverse, making accurate prediction very challenging.
In order to solve this problem, it is necessary to capture multi-scale features into FCNs.
It is well known that multi-scale features benefit many computer vision tasks such as image classification, object detection and
semantic segmentation. A straightforward approach is
to resize the input image to different scales and input each copy to
a share-parameter network. At some stage of the network, convolutional features are fused by concatenation or summation \cite{Lin2015Efficient}.
The second approach is to use skip connections to fuse features from different layers as introduced in ResNet \cite{ResNet}.
DeeplabV3 \cite{Deeplabv3} proposes the ASPP module, which applies several parallel atrous convolutions to obtain multi-scale receptive fields.
The PSPNet \cite{PSPNet} method proposes a PSP module, which applies multiple pooling with different sizes to obtain multi-scale features.
These two methods are widely used and represent the current state-of-the-art methods.
However, previous works haven't focused on how to acquire multi-scale receptive field in the backbone.
Here we propose a new \emph{multi-receptive field module (MRFM)}.
Unlike previous works, we introduce the  multi-receptive field module by re-designing the backbone work.
Meanwhile, considering the trade-off between speed and performance,
we also propose a light-weight version of MRFM, which does not introduce computation overhead for inference.

The second problem is that it is difficult to distinguish the boundaries of objects/stuff. Recently, some works
pay extra attention to the classification of pixels at edges, achieving improved results.
DFN \cite{DFN} combines the edge detection task with the semantic segmentation task.
By observing the fact that edge pixels tend to be incorrectly classified, we design a new \emph{edge aware loss}.
The main idea is that during the training, a pixel close to or on the edge is assigned a weight which incurs a weighted penalty for the classifier.

Specially, our approach achieves state-of-the-art performance on the Cityscapes dataset \cite{cityscpaes} and the Pascal VOC2012 dataset \cite{everingham2012pascal}.
Our main contributions are summarized as follows.
\begin{itemize}
    \item We propose a multi-receptive field module (MRFM), which is crucial for improving prediction performance. Meanwhile, considering the trade-off between performance and inference efficiency, we design a lightweight MRFM.
    
    \item We design an edge-aware loss to separate the boundary more accurately. The loss function does not need additional annotated data, but can be computed only on existing semantic segmentation data.
    
    \item The proposed network achieves new state-of-the-art performance on the Cityscapes and Pascal VOC2012 dataset. In particular, we achieve an mIoU score of 83.0\% on Cityscapes, and an mIoU score of 88.4\% on the Pascal VOC2012 dataset.
    
\end{itemize}

%-------------------------------------------------------------------------

\subsection{Related Works}

Models based on Fully Convolution Networks \cite{FCN1} have achieved high performance on several segmentation benchmarks\cite{everingham2012pascal,cityscpaes}. Here we review some works most relevant to ours, focusing on the issues of receptive fields and
improving edge pixel classification.

\textbf{Receptive Field:}
It is well known that the size of receptive fields is critically important for high-level computer vision tasks. In dense prediction tasks like semantic segmentation, depth estimation, it is crucial for each prediction pixel to have a receptive field that is sufficiently large such that context information can be considered for making a correct prediction. To enlarge the receptive field, the feature map's output stride is set to 8, or 16 of the input size in semantic segmentation. On the other hand, as one has to upsample the output to the original input size, a large output stride may degrade the accuracy of per-pixel prediction. That is one of the reasons why multi-scale receptive field, corresponding to multi-scale features, can be beneficial for per-pixel prediction. Another interpretation is that, to make the network robust to a certain degree of invariance to translation, FCNs employ convolutions with stride and/or pooling for this purpose. However, this translation invariance is very harmful to accurately predict the labels of pixels at the edge as a shift of several pixels may result in the same feature for pixel classification. This is dilemma. To use features from different levels of layers can partially alleviate this problem, which is essentially the same as using multiple different scales of features. In order to aggregate multi-scale context, the work of \cite{yu2015multi} employs a series of atrous convolution with increasing rates. Following \cite{yu2015multi}, deeplabv2 \cite{Deeplabv2} proposes a astrous spatial pyramid pooling (ASPP) which applies four parallel atrous convolution with different rates at the bottom of ResNet \cite{ResNet}. With the  evolution of Deeplab architectures, ASPP forms several parallel atrous convolutions and an global average pooling structure, and all of operations contain batch normalization. ASPP has shown  its  sucess in extracting features of different scales. Besides, DenseASPP \cite{DenseASPP} proposes an improved ASPP which is inspired by DenseNet \cite{DenseNet}. RefinNet \cite{RefineNet} proposes an chained residual pooling which captures background context from a large image region. PSPNet \cite{PSPNet} proposes a Spatial Pyramid Pooling (PSP) to extract feature maps with different sizes  of pooling, then concatenates all the feature maps after upsampling. Notably, all of these modules are placed at the end of the backbone. We design a different approach that achieves multi-receptive field by interleaving in the backbone.

\textbf{Objects/stuff Edge:} In the traditional image segmentation task, many methods attempt to use edge or low-level information to improve performance.
For the semantic segmentation method based on deep learning, the use of edges can be divided into two categories. The first one is using an Encoder-Decoder architecture that refines the final prediction with low-level features.
Low-level features make the results sharper. For examples, SegNet \cite{SegNet} utilizes the saved pool indices to recover the reduced spatial information. Unet \cite{Unet} uses different spatial information by skip connection. MSCI \cite{MSCI} aggregates features form different scales with two LSTM chains. The other is multi-task that shares  parameters for both semantic segmentation and edge detection, such as DFN \cite{DFN}. The work of \cite{fastermaskrcnn} shows that an auxiliary task (the edge agreement head) leads  to faster training of the mask segmentation task.

\section{Methods}
In this section, we introduce our proposed \emph{Multi-Receptive Field module} in detail first. Then, we present the \emph{Edge Aware Loss}. Finally, we describe the complete network architecture.

%-------------------------------------------------------------------------
\subsection{Multi-Receptive Field Module}

\begin{figure}[ht]
\centering
\includegraphics[scale=0.3]{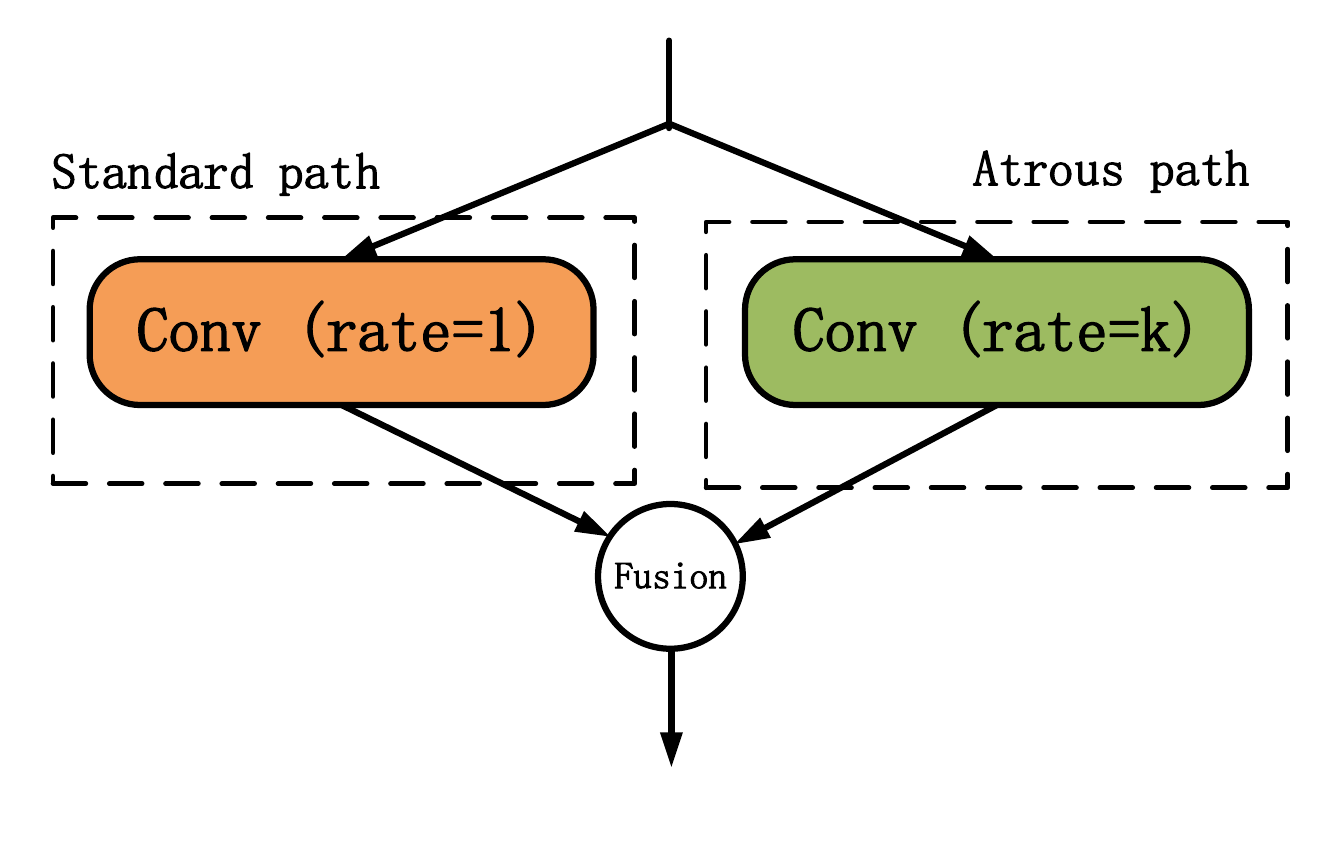}
\caption{The structure of standard Multi-Receptive Field module. Each Conv represents a basic unit in backbone, such as bottleneck in ResNet. Each path has different receptive field. The output of each path is added by the weight. Each color represents a path.}
\label{MRFMforperformancemodule}
\end{figure}

Most of the methods capture multi-receptive field by modifying the structure at the end of the network \cite{Deeplabv2,PSPNet,DenseASPP}. However, previous works have not pay much attention on how to design a backbone to capture multi-receptive field. Therefore, we propose the \emph{multi receptive field module (MRFM)} by redesigning the backbone, illustrated in the Figure \ref{MRFMforperformancemodule}.

\textbf{Multi-Receptive Field Module (MRFM):} Generally, the backbone has a basic module, such as bottleneck in ResNet \cite{ResNet}, the basic module in Xception \cite{Xception}. Take Xception for example. The basic module is composed of three depthwise convolution, one skip connection or skip connection with 1 $\times$ 1 convolution. It's easy to understand that the basic module only captures single scale receptive field. Therefore, it is difficult to segment multi-scale objects accurately. So, we propose a \emph{multi-receptive field module (MRFM)} that can simply replace any basic module, illustrated in the Figure \ref{MRFMforperformancemodule}. MRFM is composed of two paths with different receptive field. One path of MRFM is the same as the basic module, which called standard path. The other path, called atrous path, has the same structure as the basic module, but the convolution method is replaced by atrous convolution. In order to adaptively choose which receptive field is suitable, the output of each path is added by weights. In particular, two weights are normalized. To sum up, MRFM is explained by Equation \ref{MRFM_basic}, where $f_{1}(*)$ denotes the basic module in the standard path, $g_{k}(*)$ denotes the basic module replaced by atrous convolution in the atrous path ($k$ denotes dilated rate), $w_{1}$ is weights of the standard path, and $w_{2}$ is weights of the atrous path.

\begin{equation}
y= w_{1} \cdot f_{1}(x) + w_{2} \cdot g_{k}(x)
\label{MRFM_basic}
\end{equation}

\begin{figure}[ht]
\subfigure[Dual path in training stage]{
\begin{minipage}{0.4\linewidth}
\begin{center}
\includegraphics[width=\linewidth]{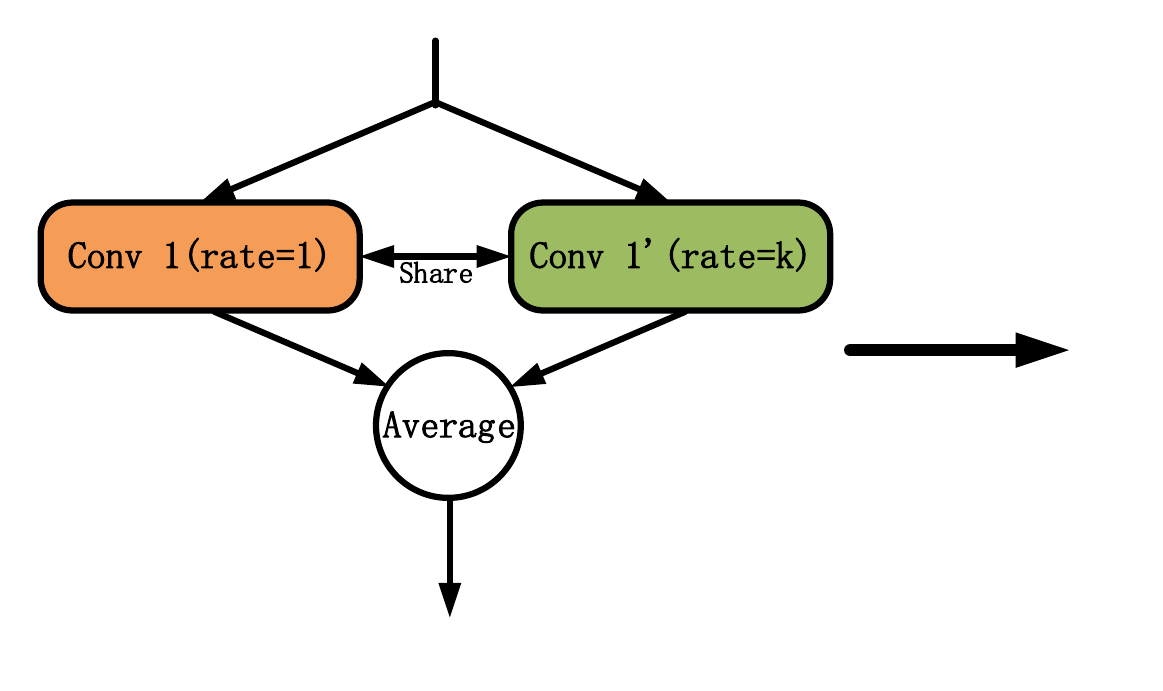}
\end{center}
\end{minipage}
\label{Dual path in training stage}
}
\subfigure[Remove atrous path]{
\begin{minipage}{0.55\linewidth}
\begin{center}
\includegraphics[width=\linewidth]{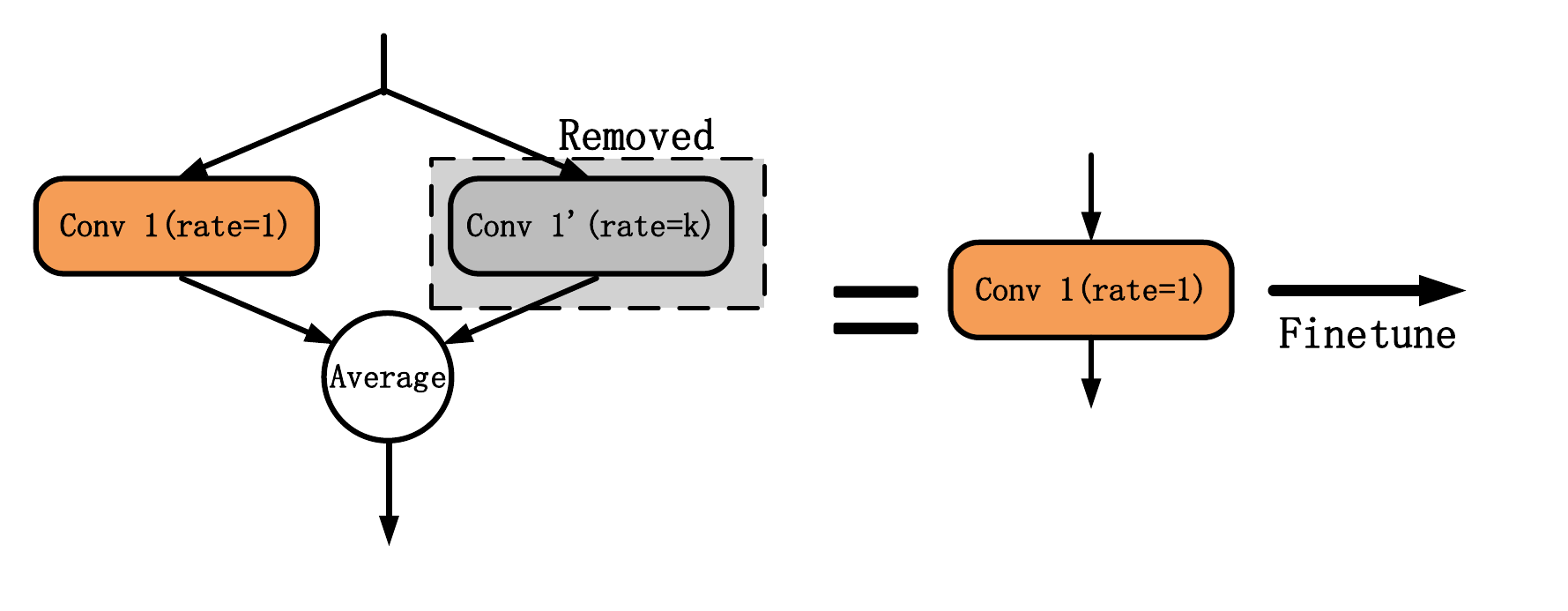}
\end{center}
\end{minipage}
\label{Remove atrous path}
}
\caption{The process of Multi-Receptive Field Module Lite. Each Conv represents a basic unit in backbone, such as bottleneck in ResNet.}
\label{MRFMforefficentmodule}
\end{figure}

\textbf{Multi-Receptive Field Module Lite(MRFM-lite):} In order to achieve a trade-off between speed and performance, we explore whether it can capture multi-scale receptive fields without increasing parameters. The two paths of MRFM are exactly the same except for the dilated rate. Now that two paths are identical except for the dilated rate, then we share all the parameters of each path. This method can effectively reduce the number of parameters, but it can still capture the multi-receptive field. Therefore, the model obtained through this structure training will have better performance. The number of parameters decrease, but the computation doesn't decrease during inference. In order to reduce the computation during inference, thanks to sharing parameters, we could remove atrous path easily, then we fine tune parameters for the better performance. In addition, when we fine tune the parameters, we use a smaller learning rate and fix the batch normalization. So, we call all of these processes Multi-Receptive Field Module Lite (MRFM-lite), illustrated in the Figure \ref{MRFMforefficentmodule}. In simple terms, the first step is to train the network with the shared parameters dual paths, and the second step is to remove atrous path and fine tune the parameters. MRFM-lite achieves a trade-off between speed and performance.

In summary, we propose two versions of MRFM. Standard MRFM could replace any basic module to capture multi-receptive field in the backbone. Although introducing parameters, MRFM performs better. MRFM-lite does not increase parameters and computation, but it still can capture multi-receptive field. MRFM and MRFM-lite both can be applied to mainstream backbone, such as ResNet, Xception.

\subsection{Edge Aware Loss}

\begin{figure}
\begin{center}
\subfigure[Image]{
\begin{minipage}{0.3\linewidth}
\begin{center}
\includegraphics[width=\linewidth]{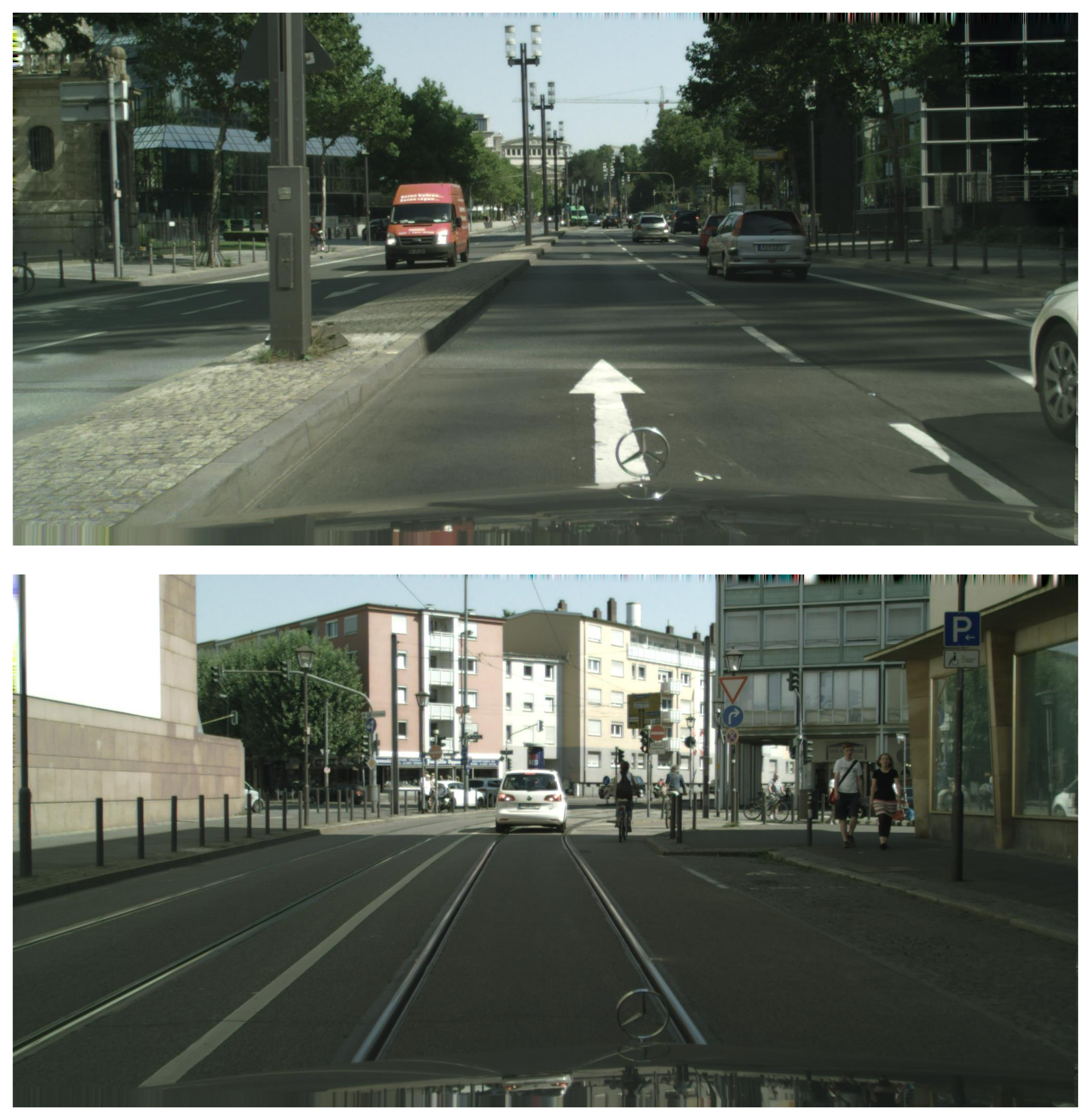}
\end{center}
\end{minipage}
\label{EAL_Image}
}
\subfigure[Ground Truth]{
\begin{minipage}{0.3\linewidth}
\begin{center}
\includegraphics[width=\linewidth]{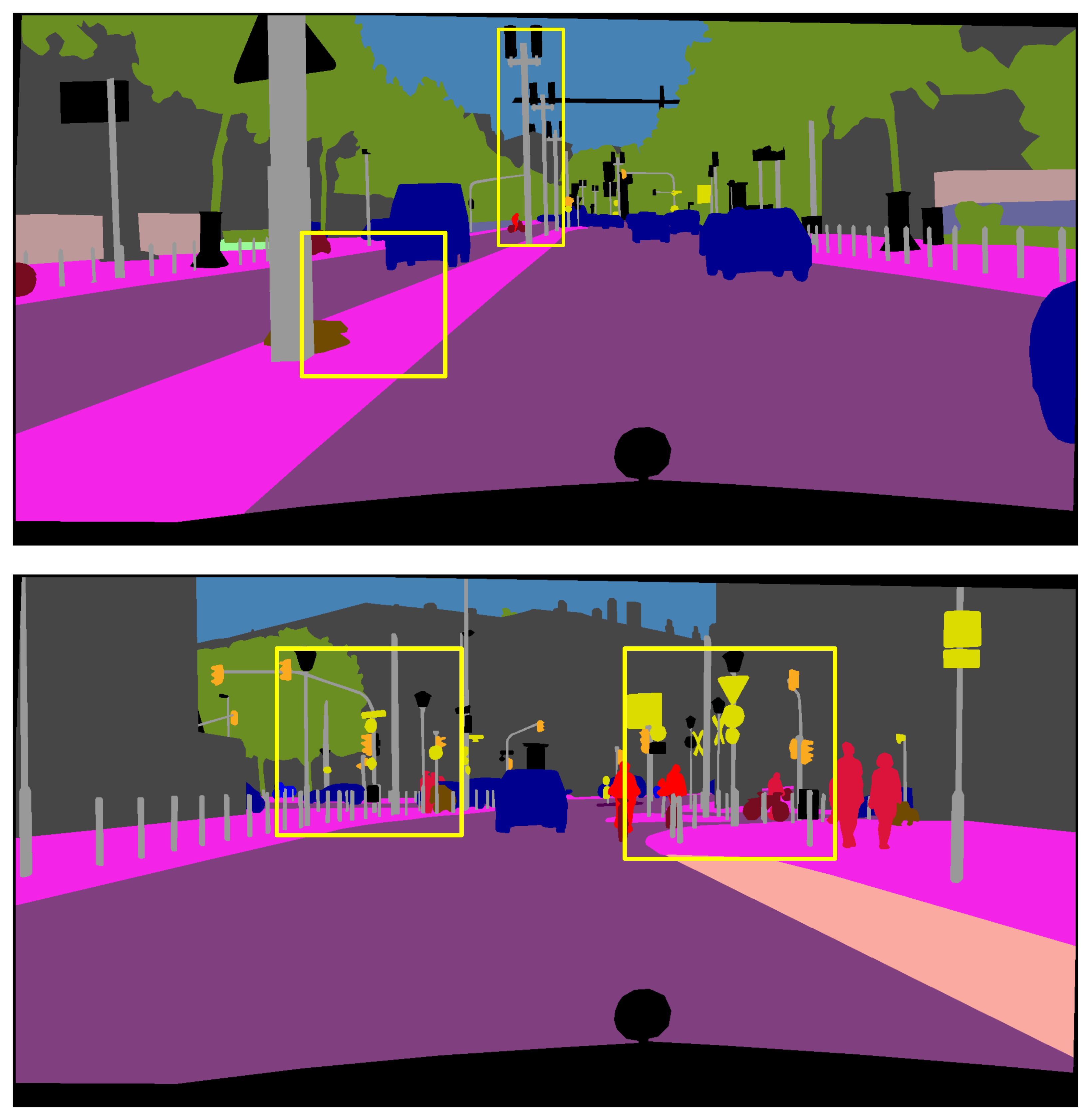}
\end{center}
\end{minipage}
\label{EAL_GT}
}
\subfigure[FCN]{
\begin{minipage}{0.3\linewidth}
\begin{center}
\includegraphics[width=\linewidth]{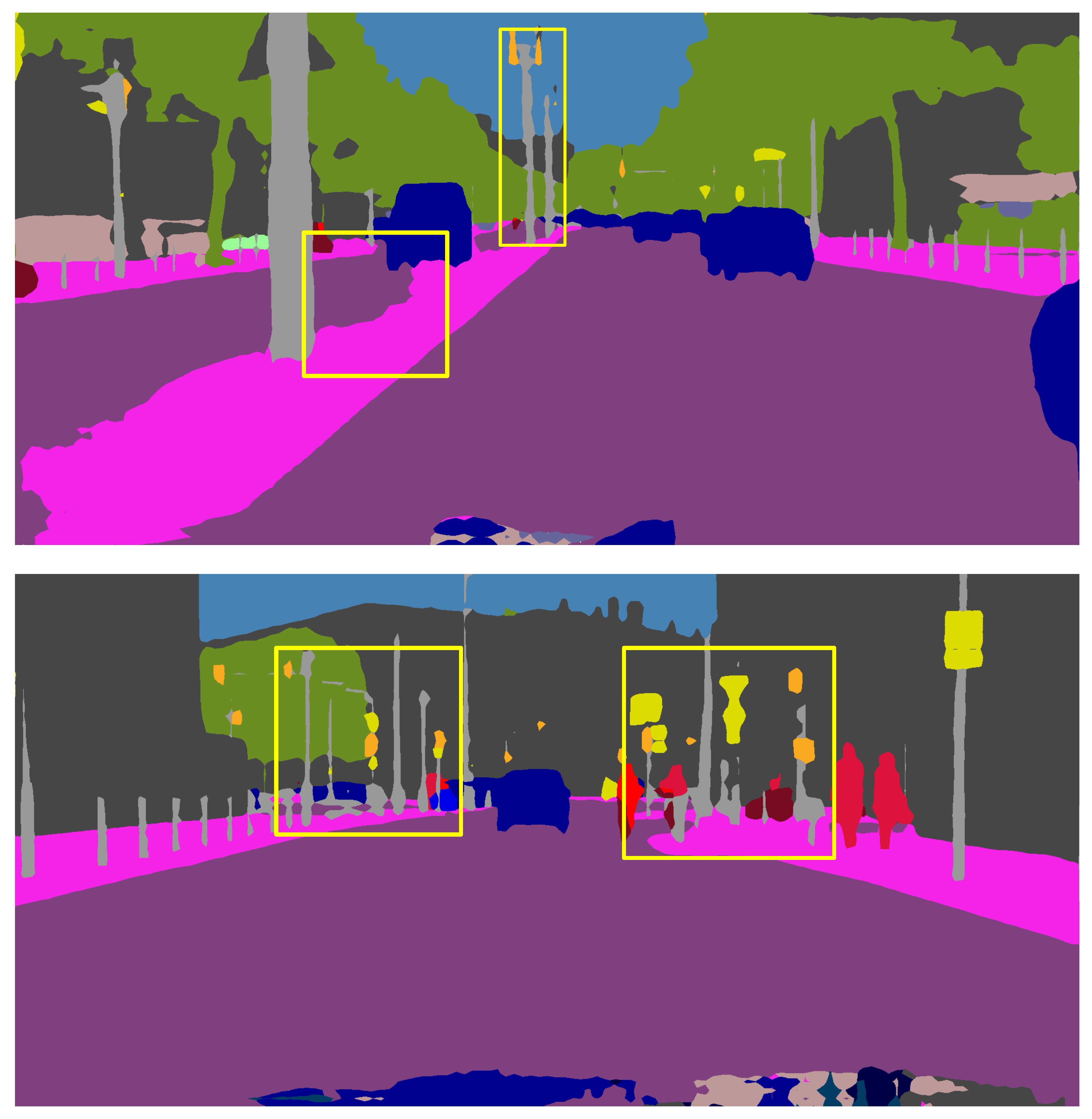}
\end{center}
\end{minipage}
\label{EAL_FCN}
}
\end{center}
\caption{ It is difficult to distinguish the boundaries of objects/stuff. Pixels located near the edge of Road and Pole are often incorrectly predicted. }
\label{EAL_error}
\end{figure}

By observing the results in the Figure \ref{EAL_error}, we find that pixels near the edge of an object often tend to be incorrectly classified. So, we propose a new edge aware loss(EAL).

\begin{figure}
\begin{center}
\includegraphics[width=\linewidth]{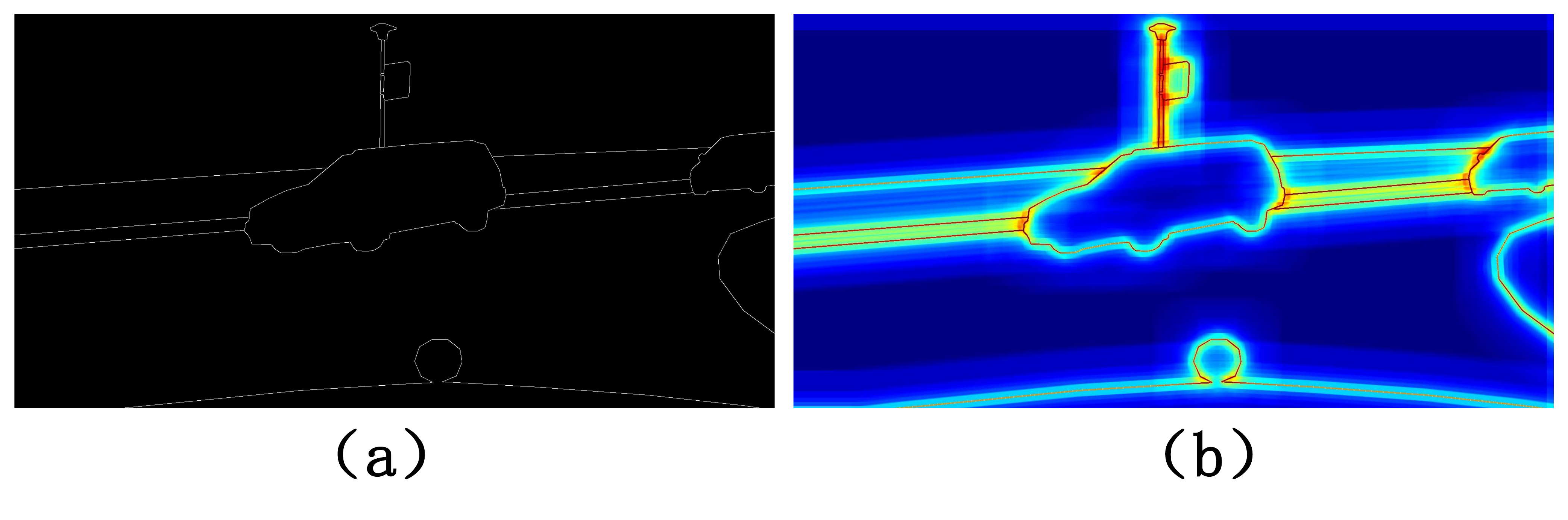}
\end{center}
\caption{The weight map of edge aware loss. (a) shows the edge detection result on the ground truth. (b) shows the weights map that edge aware loss generates. The closer the edge is, the brighter it will be.}
\label{edgewareloss}
\end{figure}

The main idea is that during the training, a pixel close to or on the edge is assigned a weight which incurs a weighted penalty for the classifier. As shown in the Figure \ref{edgewareloss}, the closer the pixel to the edge, the greater the weight. Importantly, EAL doesn't need introduce additional annotated samples. Here is the calculation process for EAL: First, edge is detected on annotated ground truth with Sobel \cite{Duda1973Pattern}, and the edge map is got. Second, to get the weight associated with the pixel's distance to the edge, we apply a k $\times$ k convolution, which is filled by 1, on the edge map. Third, to prevent the weight from getting too large, set a threshold. The entire calculation process can be expressed in the following Equation \ref{weight equation}, where $x$ denotes the ground truth, $C_{k}(*)$ donates the k $\times$ k convolution filled by 1, $E(*)$ donates the edge detection, $Threshold$ donates max value of the weight map. Finally, we get a weight map which incurs a weighted penalty for classified. This weight map multiplies the output that compute by cross entropy, as shown in Equation \ref{EAL equation}, where $y$ denotes the outputs of the network, $w$ donates the weight map.

\begin{equation}\label{weight equation}
  w = min\{C_{k}(E(x)), Threshold\}
\end{equation}
\begin{equation}\label{EAL equation}
  L = SoftmaxLoss(y;w)
\end{equation}

\begin{figure*}[ht]
\begin{center}
\includegraphics[width=0.7\linewidth]{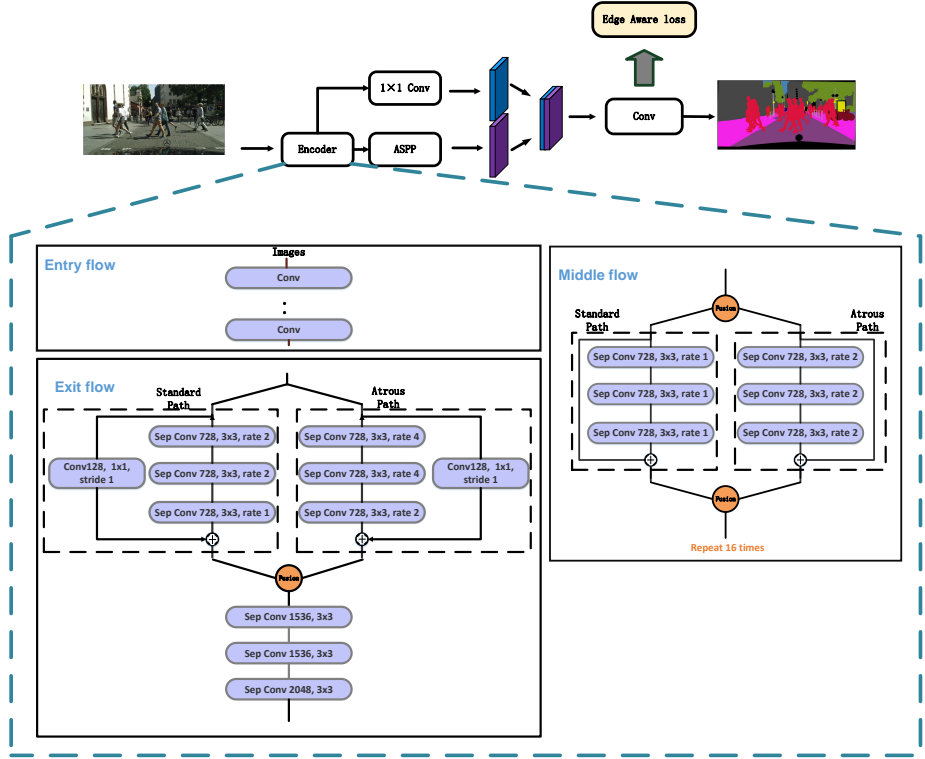}
\end{center}
\caption{Overview of Multi Receptive Field Network Architecture.}
\label{Overview of our proposed MRFM}
\end{figure*}

%-------------------------------------------------------------------------
\subsection{Network Architecture}

With the multi-receptive field module, we propose a \emph{Multi-Receptive Field Network} architecture based on Xception modified by \cite{Deeplabv3+} as illustrated in Figure \ref{Overview of our proposed MRFM} . Here we think of \cite{Deeplabv3+} as a powerful backbone. Given an input image, we use Xception model with the dilated network strategy. Meanwhile, replace the basic module in middle flow and exit flow's block1 with MRFM. In addition, the final feature map spatial size is 1/16 of the input image. Then ASPP and Decoder, which are same as \cite{Deeplabv3+}, are also employed. At last, we employ EAL to compute loss.

\section{Experiments}

We evaluate our approach on two public and widely used semantic image segmentation datasets, Cityscapes \cite{cityscpaes} and Pascal VOC2012 \cite{everingham2012pascal}. We first report the implementation details. Then we perform a series of ablation experiments on Cityscapes, and analyze the results in detail. Finally, we report our results on Pascal VOC2012 compared with other state-of-the-art methods.

%-------------------------------------------------------------------------
\subsection{Implementation Details}

Our proposed network is based on the Xception modified by \cite{Deeplabv3+} pretrained on ImageNet \cite{ImageNet}. In order to better demonstrate the improvement of the network performance of the module, our work is based on FCN.

\textbf{Training:} Following previous work \cite{Deeplabv2}, we use the "poly" learning rate policy where the base learning rate is multiplied by $(1-\frac{iter}{iter_{max}})^{power}$. The initial learning rate is set to 0.1, and power is set to 0.9. We train the network using mini-bath stochastic gradient descent(SGD) \cite{ImageNet}. Momentum and weight decay coefficients are set to 0.9 and 0.0004 respectively. The size of mini-batch is set to 12. We employ crop size of 513 during Cityscapes and Pascal VOC2012 dataset.

\textbf{Data augmentation:} Following \cite{Deeplabv3}, we use mean subtraction, apply data augmentation by randomly scaling the input images with 7 scales $\{0.5, 0.75, 1, 1.25, 1.5, 1.75, 2.0\}$ and randomly left-right flipping during training.

%-------------------------------------------------------------------------
\subsection{Cityscapes Dataset}

\textbf{Dataset and Evaluation Metrics: } Cityscapes is a large and famous city street scene semantic segmentation dataset \cite{cityscpaes}. 19 classes of which 30 classes of this dataset are considered for training and evaluation. Each image has a high resolution 2048 $\times$ 1024 pixels. Cityscapes has 5,000 images taken from 50 different cities which are all fine annotation. In these 5,000 images, there are 2,979 images for training, 500 images for validation and 1,525 images for testing. There are also 19,998 images with coarse annotation. Cityscapes In the same image, cityscapes have many scale objects, such as road, sky, vehicles, persons and so on. In the same category, persons, cars, sky and so on also have a lot of scales. So cityscapes is a dataset that is extremely sensitive to the scale. In this challenging dataset, our method outperforms previous methods again. For evaluation, \emph{mean class-wise intersection over union} (mIoU) is used.

\begin{table}[t]
\begin{center}
\resizebox{0.9\linewidth}{!}{
\begin{tabular}{l c}
\hlineB{3}
Method & mIoU(\%) \\
\hline\hline
Xception41-Baseline                                 & 72.5 \\
Xception65-Baseline                                 & 73.1 \\
Xception71-Baseline                                 & 74.0 \\
Xception41+ASPP                                     & 75.0 \\
Xception41+MRFM-lite-2(DP)                          & 74.7 \\
Xception41+MRFM-lite-2(OP)                          & {\bf75.2} \\
Xception41+MRFM-lite-4(OP)                          & 72.2 \\
Xception41+MRFM-lite-4(DP)                          & 72.4 \\
Xception41+MRFM-2                                   & 75.7 \\
Xception41+MRFM-4                                   & {\bf75.9} \\
Xception41+MRFM-8                                   & 72.3 \\
Xception41+MRFM-4(Exit flow)                        & 74.3 \\
Xception41+MRFM-4(Middle flow)                      & 74.2 \\
\hlineB{3}
\end{tabular}
}
\end{center}
\caption{Investigation of MRFM with different settings. Baseline is Xception41-based FCN in which output stride is 16. MRFM-(*) present dilated rate in the atrous path. DP present dual path of MRFM-lite in training stage. OP present MRFM-lite is fine tuned with one path. MRFM(Exit flow) is represented as replacing the exit flow module with MRFM module. MRFM(Middle flow) is represented as replacing the middle flow module with MRFM module. To simplify the presentation, MRFM without special instructions replaces the basic module in middle flow and exit flow.}
\label{XceptionwithMRFM}
\end{table}

\begin{figure*}[ht]
\begin{center}
\subfigure[Image]{
\includegraphics[width=0.18\linewidth]{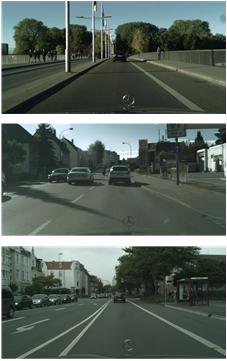}
}
\subfigure[Ground Truth]{
\includegraphics[width=0.18\linewidth]{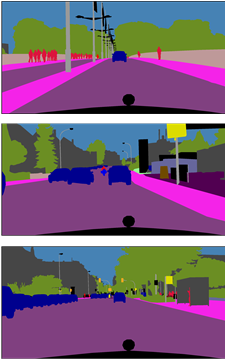}
}
\subfigure[ASPP]{\includegraphics[width=0.18\linewidth]{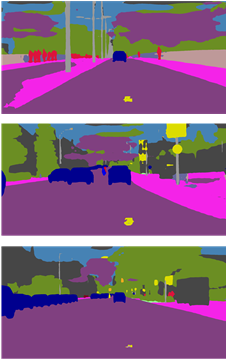}}
\subfigure[MRFM-lite]{\includegraphics[width=0.18\linewidth]{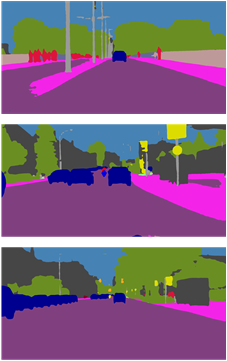}}
\subfigure[MRFM]{\includegraphics[width=0.18\linewidth]{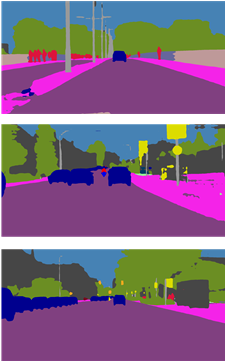}}
\end{center}
\caption{Visual improvements on Cityscapes. Based on FCN, MRFM and MRFM-lite produces more accurate and detailed results.}
\label{MRFI Result compared figure}
\end{figure*}

%-------------------------------------------------------------------------
\textbf{Ablation Study for MRFM in Xception: } In order to evaluate the improvement of MRFM and MRFM-lite, we carry out some experiments. As listed in the Table \ref{XceptionwithMRFM}, without more parameters, MRFM-lite-2 significantly improves the segmentation performance over the Xception41-baseline model by 2.7\%. MRFM-lite is trained in two stages, and we also provide the results of two stages. Compared with baseline, MRFM-lite-2(DP) get a 2.2\% promotion. Compared with MRFM-lite-2(DP), MRFM-lite-2(OP), which is fine tuned with one path, get a 0.5\% promotion. Particularly, when we fine tune MRFM-lite, we set the learning rate to 0.01 and fix batch normalization. We find it is important to fix batch normalization. It is worth noting that compared with Xception41, MRFM-lite has the same structure, the same speed and higher performance. In order to fairly compare the performance of MRFM under the similar model capacity, we also give the baseline performance of Xception65 and Xception71. With the increase of model capacity, the performance of the model has been improved, but MRFM is still better than Xcpetion71, which has the largest model capacity. Specially, with MRFM-4, the network is the best, which increases 3.4\% that is compared with Xception41, and increases 1.9\% that is compared with Xception71! In addition, compared with ASPP that is famous method to capture multi-scale receptive field, MRFM-lite is slightly better than ASPP without introducing more parameters and time, and MRFM also has a great improvement over ASPP.

\begin{figure}[t]
\begin{center}
\includegraphics[width=\linewidth]{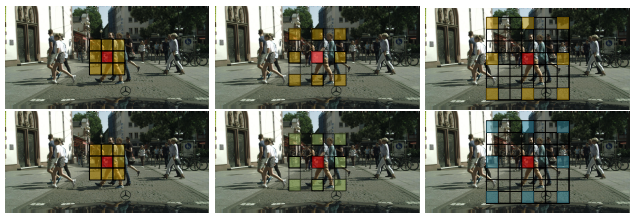}
\end{center}
\caption{Different receptive field. The same color represents the same parameters. The top row shows MRFM-lite. The bottom row shows standard MRFM. Each row shows different MRFM's receptive field. As shown in each row, receptive field becomes large different by increasing dilation rate.}
\label{receptive fields}
\end{figure}

The receptive fields of both paths interact with each other. So MRFM-lite-2 is better than MRFM-lite-4. However, in the standard MRFM, MRFM-4 is better than MRFM-2, but MRFM-8 is very terrible. As shown in Figure \ref{receptive fields}, with the increase of the dilated rate, the information of different receptive field becomes more different. So, if the receptive field of each path is much different, it is harmful. In addition, the convolution parameters of the two paths in MRFM-lite are more relevant than those of the two paths in MRFM. Therefore, the difference of dilated rate that is in the each path of MRFM can be greater. It is precisely because the receptive fields are not mutually constrained that MRFM is 0.7\% higher than MRFM-lite.

Comparing the performance differences of different parts using MRFM. MRFM in exit flow and MRFM in middle flow both bring about improvement than baseline. MRFM in both middle flow and exit flow is better than MRFM in one flow.

At last, all of experiments show that we use MRFM when we need higher accuracy and MRFM-lite when we need trade-off speed and accuracy. Both versions of MRFM can improve the basic performance of the model.

\textbf{Ablation Study for MRFM in ResNet: } In order to demonstrate the general of MRFM, we also product some experiments on ResNet50. As illustrated in the Table \ref{ResNetwithMRFM}, all of settings are better than baseline. MRFM-4 has increased 4.8\% over the baseline network. Compared with FCN, MRFM-lite-2 has improved 3.5\% without any more parameters. Compared with MRFM-lite-2(DP), MRFM-lite-2(OP) has improved 1.8\% . These experiments show MRFM and MRFM-lite could extend to other backbone easily. In addition, compared with ASPP, MRFM and MRFM-lite are also standout.Through Xception and Resnet experiments, we can boldly speculate that MRFM can be applied to other backbone.

\begin{table}[htbp]
\begin{center}
\resizebox{0.8\linewidth}{!}{
\begin{tabular}{l c}
\hlineB{3}
Method & mIoU(\%) \\
\hline\hline
ResNet50-Baseline         & 65.9 \\
ResNet50+ASPP             & 69.3 \\
ResNet50+MRFM-lite-2(DP)  & 67.2 \\
ResNet50+MRFM-lite-2(OP)  & 69.4 \\
ResNet50+MRFM-4           & {\bf 70.7} \\
\hlineB{3}
\end{tabular}
}
\end{center}
\caption{In order to demonstrate the generalization ability of the MRFM better, ResNet experiments are conduct}
\label{ResNetwithMRFM}
\vspace{-1em}
\end{table}

\textbf{Ablation Study for Receptive Field: } As illustrated in the Table \ref{AbalationMRFMOnResNet}, some experiments are organized to represent the effects of different receptive field methods. InceptionV4 \cite{szegedy2017inception} uses a lot of pooling operations, and the operation of pooling will result in the loss of details. So the result of InceptionV4 \cite{szegedy2017inception} is very poor. ResNet50-DCN \cite{zhu2019deformable} applies deformable convolution in the stage2, stage3 and stage4. DCN \cite{zhu2019deformable} effectively improved performance by 2.6\%. However, MRFM-lite is still 0.9\% better than ResNet50-DCN, and MRFM is still 2.2\% better than ResNet50-DCN. We think that DCN is an implicit way to learn multi-scale information, but MRFM and MRFM-lite are an explicit way to learn multi-scale information, so MRFM and MRFM-lite perform better.

\begin{table}[htbp]
\begin{center}
\resizebox{0.7\linewidth}{!}{
\begin{tabular}{l c}
\hlineB{3}
Method & mIoU(\%) \\
\hline\hline
InceptionV4           & 51.5 \\
ResNet50              & 65.9 \\
ResNet50-DCN          & 68.5 \\
ResNet50+MRFM-lite-2  & 69.4 \\
ResNet50+MRFM-4       & 70.7 \\
\hlineB{3}
\end{tabular}
}
\end{center}
\caption{Much work has been done to obtain different scales of the receptive field. The experiments are conduct. DCN present the deformable convolution \cite{zhu2019deformable}. The output stride of InceptionV4 \cite{szegedy2017inception} is 16.}
\label{AbalationMRFMOnResNet}
\vspace{-1em}
\end{table}

\begin{table}[t]
\begin{center}
\resizebox{0.6\linewidth}{!}{
\begin{tabular}{l c c c}
\hlineB{3}
Method & mIoU(\%) \\
\hline\hline
Cross Entropy  & 72.5 \\
Edge Branch    & 72.7 \\
OHEM \cite{Shrivastava2016Training, Wu2016Wider}          & 73.0 \\
Lovas Loss \cite{berman2018lovasz}    & 72.8 \\
EAL            & 74.1 \\
\hlineB{3}
\end{tabular}
}
\end{center}
\caption{Different loss functions brings different improvement. To prove EAL's more efficient, some experiments are conduct. }
\label{AblationOnEAL}
\vspace{-1em}
\end{table}

\begin{figure}[t]
\begin{center}
\includegraphics[width=\linewidth]{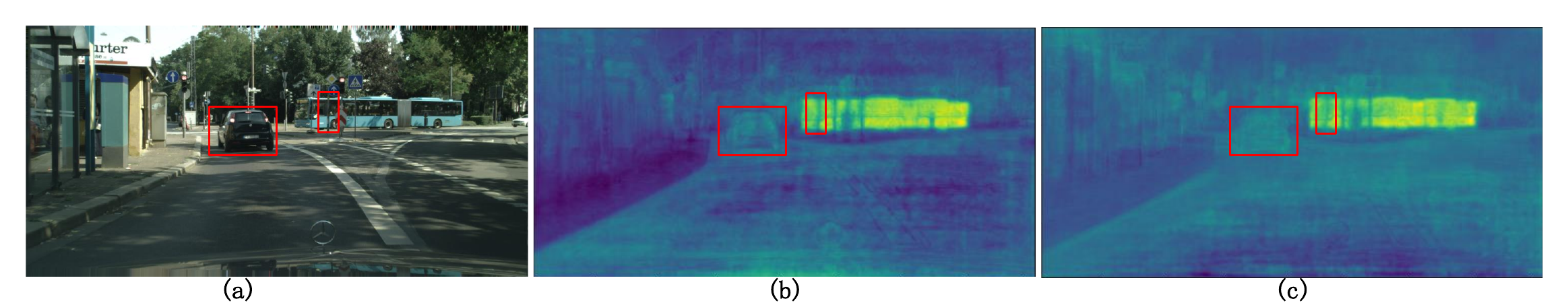}
\end{center}
\caption{Visual improvements with EAL. (a) shows RGB image; (b) shows feature map with EAL; (c) shows feature map without EAL.}
\label{edge compared image}
\vspace{-1em}
\end{figure}

%-------------------------------------------------------------------------
\textbf{Ablation Study for Edge Aware Loss: } The introduced EAL helps to detect object contour while not affecting learning process in the main master. In particular, as we have mentioned before, EAL doesn't require additional annotations. In the Table \ref{AblationOnEAL}, compared with baseline, EAL brings 1.6\% improvement. DFN \cite{DFN} proposed the multi task methods, including edge segmentation task and semantic segmentation task. So EAL is also compared with edge branch method. When adding edge branch, the model shares the backbone, and each task has its own two convolution operations. As illustrated in the Table \ref{AblationOnEAL}, the result is almost similar with Cross Entropy. Because different tasks have its own unique parameters, It's hard for edge branch to influence the main task. In addition, edges can be considered as difficult samples, so we also compare OHEM \cite{Shrivastava2016Training, Wu2016Wider}. Compared with OHEM \cite{Shrivastava2016Training, Wu2016Wider}, EAL brings 1.1\% improvement. Lovas loss solves the gap between loss function and mIoU in semantic segmentation. To better illustrate the effectiveness of EAL, we also compared the difference with Lovas loss. Compared with Lovas loss, EAL brings 1.3\% improvement. In the Figure \ref{edge compared image}, with EAL, edges can be clearly seen. When using EAL, two hyperparameters are added. One is k, which determines the longest distance of the edge, and the other is m, which determines the maximum weight. In the Table \ref{EALtabel}, it shows that EAL can improve steadily under different hyperparameters.

\begin{table}[t]
\begin{center}
\resizebox{0.5\linewidth}{!}{
\begin{tabular}{l c c c}
\hlineB{3}
Method & k & m & mIoU(\%) \\
\hline\hline
Baseline  &   &        & 78.1 \\
EAL       & 5 & 3      & 78.6 \\
EAL       & 7 & 3      & 78.5 \\
\hlineB{3}
\end{tabular}
}
\end{center}
\caption{In order to demonstrate the influence of the different hyperparameters, experiments are conduct. Except that backbone is replaced with Xception41, the rest of baseline is the same as deeplabv3+.}
\label{EALtabel}
\vspace{-1em}
\end{table}

\textbf{Framework Details: }In order to show the effect of each module in our framework and show the continuous improvement with our methods, we conducted some experiments, illustrated in the Table \ref{Ablation Study for the each module}. Based on FCN (Xception41) and deeplabv2 (Xception41+ASPP) \cite{Deeplabv2}, MRFM, MRFM-lite and EAL could bring improvement. Based on powerful deeplabv3+ (Xception65) \cite{Deeplabv3+}, our MRFM could bring 1.4\%. It is noteworthy that MRFM has continued to improve with ASPP. Although MRFM has acquired multi-scale information in backbone, as shown in the Table \ref{Ablation Study for the each module}, ASPP can still obtain multi-scale information. Since ASPP is located behind backbone and is closer to output, it is meaningful to obtain multi-scale in any part of the network. We can also see that MRFM does improve the performance of backbone, so MRFM does have the prospect of improving segmentation baseline. In addition, we speculate that replacing ASPP with better modules will lead to better results, such as replacing DPC \cite{chen2018searching}, DRN \cite{zhuang2018dense}, etc.
The highest performance MRFM can be improved by 0.5\% with EAL and get a mIoU of 80.2\% on the validation. At last, we get the best performance model, which contains deeplabv3+, MRFM-4 and EAL.

\begin{table}[t]
\begin{center}
\resizebox{\linewidth}{!}{
\begin{tabular}{l c c c c c}
\hlineB{3}
Baseline                          & ASPP             & MRFM-lite-2 &  MRFM-4       &EAL          & mIoU \\
\hline\hline
Xception41                        &                  &             &                    &             & 72.5 \\
Xception41                        &                  &$\checkmark$ &                    &             & 75.2 \\
Xception41                        &                  &             &  $\checkmark$      &             & 75.9 \\
Xception41                        & $\checkmark$     &             &                    &             & 75.0 \\
Xception41                        & $\checkmark$     &$\checkmark$ &                    &             & 76.2 \\
Xception41                        & $\checkmark$     &$\checkmark$ &                    &$\checkmark$ & 76.3 \\
Xception41                        & $\checkmark$     &             & $\checkmark$       &             & 76.7 \\
Xception41                        & $\checkmark$     &             & $\checkmark$       &$\checkmark$ & 77.7 \\
\hline\hline
Xception65+Decoder                & $\checkmark$     &             &                    &             & 78.4 \\
Xception65+Decoder                & $\checkmark$     &             & $\checkmark$       &             & 79.7 \\
Xception65+Decoder                & $\checkmark$     &             & $\checkmark$       &$\checkmark$ & {\bf 80.2} \\

\hlineB{3}
\end{tabular}
}
\end{center}
\caption{Ablation Study for the each module.}
\label{Ablation Study for the each module}
\end{table}

\textbf{Ablation Study for different scale testing:} In order to show more clearly that MRFM performs better than single-scale model in different scales, we organized a group of experiments. Image sent to the network are scaled to form different scales. Illustrated in the Tabel \ref{differentscaleresult}, the results of each scale are more better than the results of single-scale model. Moreover, by calculating the standard deviation, it is found that the results fluctuate less in each scale, so it can be proved that MRFM is robust for multi-scale.

\begin{table}[tbhp]
\begin{center}
\resizebox{\linewidth}{!}{
\begin{tabular}{l c c c c c c c |c}
\hlineB{3}
Method     & 0.5s & 0.75s& 1.0s& 1.25s& 1.5s& 1.75s  & 2.0s  & STD \\
\hline\hline
Baseline   & 72.7 & 76.8 & 78.4 & 78.8 & 78.6 & 77.9 & 76.6  & 1.97\\
MRFM       & 75.1 & 78.4 & 79.8 & 80.3 & 80.0 & 79.6 & 78.1  & 1.68\\
\hlineB{3}
\end{tabular}
}
\end{center}
\caption{Each column represents how many times the input image is the original image. The results show that MRFM is superior to single-scale model in different scales. The baseline is same as Deeplabv3+.}
\label{differentscaleresult}
\vspace{-1em}
\end{table}

\begin{table*}[t]
\begin{flushleft}
\begin{center}
\resizebox{\textwidth}{!}{
\begin{tabular}{l|ccccccccccccccccccc|c}
\hlineB{3}
\toprule
\textbf{Methods}  & \textbf{road} & \textbf{swalk} & \textbf{build.} & \textbf{wall} & \textbf{fence} & \textbf{pole} & \textbf{tlight} & \textbf{sign} & \textbf{veg} & \textbf{terrain} & \textbf{sky} & \textbf{person} & \textbf{rider} & \textbf{car} & \textbf{truck} & \textbf{bus} & \textbf{train} & \textbf{mbike} & \textbf{bike} & \textbf{IoU}\\
\hline
DUC \cite{Wang2017Understanding}     & 98.5  & 85.9  & 93.2  & 57.7  & 61.1  & 67.2  & 73.7  & 78.0  & 93.4  & 72.3  & 95.4  & 85.9  & 70.5  & 95.9  & 76.1  & 90.6  & 83.7  & 67.4  & 75.7  & 80.1\\
DFN \cite{DFN}     & 98.6  & 85.9  & 93.2  & 59.6  & 61.0  & 66.6  & 73.2  & 78.2  & 93.5  & 71.6  & 95.5  & 86.5  & 70.5  & 96.1  & 77.1  & 89.9  & 84.7  & 68.2  & 76.5  & 80.3\\
ResNet38 \cite{Wu2016Wider}   & 98.7  & 86.9  & 93.3  & 60.4  & 62.9  & 67.6  & 75.0  & 78.7  & 93.7  & 73.7  & 95.5  & 86.8  & 71.1  & 96.1  & 75.2  & 87.6  & 81.9  & 69.8  & 76.7  & 80.6\\
PSPNet \cite{PSPNet}   & 98.7  & 86.9  & 93.5  & 58.4  & 63.7  & 67.7  & 76.1  & 80.5  & 93.6  & 72.2  & 95.3  & 86.8  & 71.9  & 96.2  & 77.7  & 91.5  & 83.6  & 70.8  & 77.5  & 81.2\\
Deeplabv3 \cite{Deeplabv3}   & 98.6  & 86.2  & 93.5  & 55.2  & 63.2  & 70.0  & 77.1  & 81.3  & 93.8  & 72.3  & 95.9  & 87.6  & 73.4  & 96.3  & 75.1  & 90.4  & 85.1  & 72.1  & 78.3  & 81.3\\
AdapNet++ \cite{AdapNet++} &98.6 &86.2 &93.3 &57.8 &62.0 &67.3 &75.0 &79.6 &93.6 &72.3 &95.3 &86.4 &72.2 &96.2 &81.5 &92.4 &88.0 &71.2 &76.6 &81.3 \\
Mapillary \cite{Bul2017In}   & 98.4  & 85.0  & 93.6  & 61.7  & 63.9  & 67.7  & 77.4  & 80.8  & 93.7  & 71.9  & 95.6  & 86.7  & 72.8  & 95.7  & 79.9  & 93.1  & 89.7  & 72.6  & 78.2  & 82.0\\
Deeplabv3+ \cite{Deeplabv3+}  & 98.7  & 87.0  & 93.9  & 59.5  & 63.7  & 71.4 & 78.2  & 82.2\cellcolor{gray!40}  & 94.0 & 73.0  & 95.8  & 88.0  & 73.3  & 96.4  & 78.0  & 90.9  & 83.9  & 73.8  & 78.9  & 82.1 \\
AutoDeeplab \cite{Liu2019Auto} &98.8\cellcolor{gray!40} &87.6 	&93.8 	&61.4 	&64.4 	&71.2 	&77.6 	&80.9 	&94.1 	&72.7 	&96.0\cellcolor{gray!40} 	&87.8 	&72.8 	&96.5\cellcolor{gray!40} 	&78.2 	&90.9 	&88.4 	&69.0 	&77.6 &82.1\\
DPC \cite{chen2018searching}     & 98.7  & 87.1  & 93.8  & 57.7  & 63.5  & 71.0  & 78.0  & 82.1  & 94.0 & 73.3  & 95.4  & 88.2  & 74.5  & 96.5 \cellcolor{gray!40} & 81.2\cellcolor{gray!40}  & 93.3\cellcolor{gray!40}  & 89.0 \cellcolor{gray!40} & 74.1\cellcolor{gray!40}  & 79.0 \cellcolor{gray!40} & 82.7 \\
DRN \cite{zhuang2018dense}  & 98.8\cellcolor{gray!40}  & 87.7   & 94.0    & 65.1\cellcolor{gray!40}        & 64.2          & 70.1       & 77.4        & 81.6         & 93.9        & 73.5         & 95.8          & 88.0          & 74.9\cellcolor{gray!40}       & 96.5\cellcolor{gray!40}                    & 80.8      & 92.1      & 88.5         & 72.1           & 78.8            & 82.8             \\
\hline
{\bf Ours}   &98.8\cellcolor{gray!40} 	&88.0\cellcolor{gray!40} 	&94.2\cellcolor{gray!40} 	&63.8 	&64.7\cellcolor{gray!40} 	&72.3\cellcolor{gray!40} 	&78.3\cellcolor{gray!40} 	&81.8 	&94.2 \cellcolor{gray!40} 	&73.9\cellcolor{gray!40} 	&95.7 	&88.3\cellcolor{gray!40} 	&74.6 	&96.4 	&79.5 	&92.2 	&88.1 	&72.8 	&78.6 &83.0\cellcolor{gray!40} \\

\bottomrule
\hlineB{3}
\end{tabular}
}
\end{center}
\end{flushleft}
\caption{Results on Cityscapes testing set. Methods are trained using both fine and coarse data. The best entry in each columns is marked in gray color.(Note: the methods that only use cityscapes dataset are included.)}
\label{Results on test}
\end{table*}

\begin{table*}[t]
\begin{flushleft}
\resizebox{\textwidth}{!}{
\begin{tabular}{l|cccccccccccccccccccc|c}
\hlineB{3}
\toprule
\textbf{Method}  & \textbf{aero} & \textbf{bike} & \textbf{bird} & \textbf{boat} & \textbf{bottle} & \textbf{bus} & \textbf{car} & \textbf{cat} & \textbf{chair} & \textbf{cow} & \textbf{table} & \textbf{dog} & \textbf{horse} & \textbf{mbike} & \textbf{person} & \textbf{plant} & \textbf{sheep} & \textbf{sofa} & \textbf{train} & \textbf{tv} & \textbf{mIoU}\\
\hline
RefineNet \cite{RefineNet}   & 95.0  & 73.2  & 93.5  & 78.1  & 84.8  & 95.6  & 89.8  & 94.1  & 43.7  & 92.0  & 77.2  & 90.8  & 93.4  & 88.6  & 88.1  & 70.1  & 92.9  & 64.3  & 87.7  & 78.8  & 84.2\\
ResNet38 \cite{Wu2016Wider}   & 96.2  & 75.2  & 95.4  & 74.4  & 81.7  & 93.7  & 89.9  & 92.5  & 48.2  & 92.0  & 79.9  & 90.1  & 95.5  & 91.8  & 91.2  & 73.0  & 90.5  & 65.4  & 88.7  & 80.6  & 84.9\\
PSPNet \cite{PSPNet}   & 95.8  & 72.7  & 95.0  & 78.9  & 84.4  & 94.7  & 92.0  & 95.7  & 43.1  & 91.0  & 80.3  & 91.3  & 96.3  & 92.3  & 90.1  & 71.5  & 94.4  & 66.9  & 88.8  & 82.0  & 85.4\\
Deeplabv3 \cite{Deeplabv3}   & 96.4  & 76.6  & 92.7  & 77.8  & 87.6  & 96.7  & 90.2  & 95.4  & 47.5  & 93.4  & 76.3  & 91.4  & 97.2\cellcolor{gray!40}  & 91.0  & 92.1  & 71.3  & 90.9  & 68.9  & 90.8  & 79.3  & 85.7\\
EncNet \cite{zhang2018context}   & 95.3  & 76.9  & 94.2  & 80.2  & 85.3  & 96.5  & 90.8  & 96.3  & 47.9  & 93.9  & 80.0  & 92.4  & 96.6  & 90.5  & 91.5  & 70.9  & 93.6  & 66.5  & 87.7  & 80.8  & 85.9\\
DFN \cite{DFN}     & 96.4  & 78.6  & 95.5  & 79.1  & 86.4  & 97.1  & 91.4  & 95.0  & 47.7  & 92.9  & 77.2  & 91.0  & 96.7  & 92.2  & 91.7  & 76.5  & 93.1  & 64.4  & 88.3  & 81.2  & 86.2\\
SDN \cite{fu2017stacked}     & 96.9  & 78.6  & 96.0  & 79.6  & 84.1  & 97.1  & 91.9  & 96.6  & 48.5  & 94.3  & 78.9  & 93.6  & 95.5  & 92.1  & 91.1  & 75.0  & 93.8  & 64.8  & 89.0  & 84.6  & 86.6\\
Deeplabv3+ \cite{Deeplabv3+}   & 97.0  & 77.1  & 97.1\cellcolor{gray!40}  & 79.3  & 89.3  & 97.4\cellcolor{gray!40}  & 93.2  & 96.6  & 56.9  & 95.0  & 79.2  & 93.1  & 97.0  & 94.0  & 92.8  & 71.3  & 92.9  & 72.4  & 91.0  & 84.9  & 87.8\\
ExFuse \cite{zhang2018exfuse}   & 96.8  & 80.3 \cellcolor{gray!40}  & 97.0  & 82.5\cellcolor{gray!40}  & 87.8  & 96.3  & 92.6  & 96.4  & 53.3  & 94.3  & 78.4  & 94.1\cellcolor{gray!40}  & 94.9  & 91.6  & 92.3  & 81.7\cellcolor{gray!40}  & 94.8\cellcolor{gray!40}  & 70.3  & 90.1  & 83.8  & 87.9\\
MSCI \cite{MSCI}    & 96.8  & 76.8  & 97.0  & 80.6  & 89.3  & 97.4\cellcolor{gray!40}  & 93.8\cellcolor{gray!40}  & 97.1\cellcolor{gray!40}  & 56.7  & 94.3  & 78.3  & 93.5  & 97.1  & 94.0  & 92.8  & 72.3  & 92.6  & 73.6  & 90.8  & 85.4 \cellcolor{gray!40} & 88.0\\
\hline
{\bf Ours} & 97.1\cellcolor{gray!40} &78.6 &97.1\cellcolor{gray!40} &80.6 &89.7\cellcolor{gray!40} &97.3 &93.6 &96.7 &59.0\cellcolor{gray!40} &95.4\cellcolor{gray!40} &81.1\cellcolor{gray!40} &93.2 &97.5\cellcolor{gray!40} &94.2\cellcolor{gray!40} &92.9\cellcolor{gray!40} &72.3 &93.1 &74.2 &91.0 &85.0 &88.4\cellcolor{gray!40}\\
\bottomrule
\hlineB{3}
\end{tabular}
}
\end{flushleft}
\caption{Per-class results on PASCAL VOC2012 testing set. Methods are all pre-trained on MS-COCO. The best entry in each columns is marked in gray color.(Note: the methods that use the public dataset are included.)}
\label{pascalvoctest}%

\end{table*}%

\begin{figure}[t]
\begin{center}
\includegraphics[width=\linewidth]{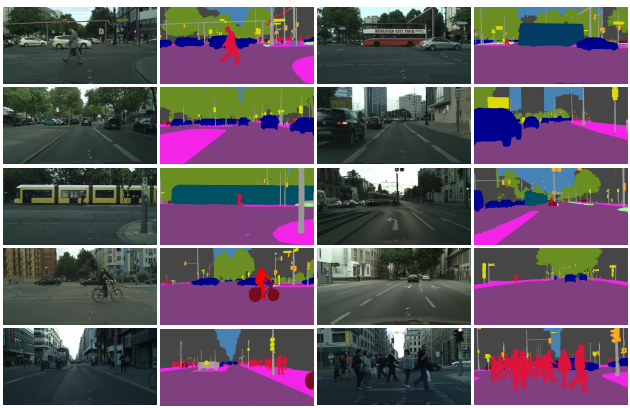}
\end{center}
\caption{Visualization results on Cityscapes test set.}
\label{Cityscapes test set visual}
\end{figure}

\textbf{Comparing with State-of-the-art: }
We further compare our method with the state-of-the-art methods on the Cityscapes test set. For maximum performance, we use the best structure in the Table \ref{Ablation Study for the each module}, which consists of DeeplabV3+(Xception65), MRFM and EAL. In particular, we use only fine-coarse data training networks, and submit our results to the official evaluation server. Results are shown in Table \ref{Results on test}. The result shows that our method is better than previous method in some categories, such as building, pole, traffic light, terrain, person. Because these classes have diverse scales.

\begin{figure}[ht]
\begin{center}
\includegraphics[width=\linewidth]{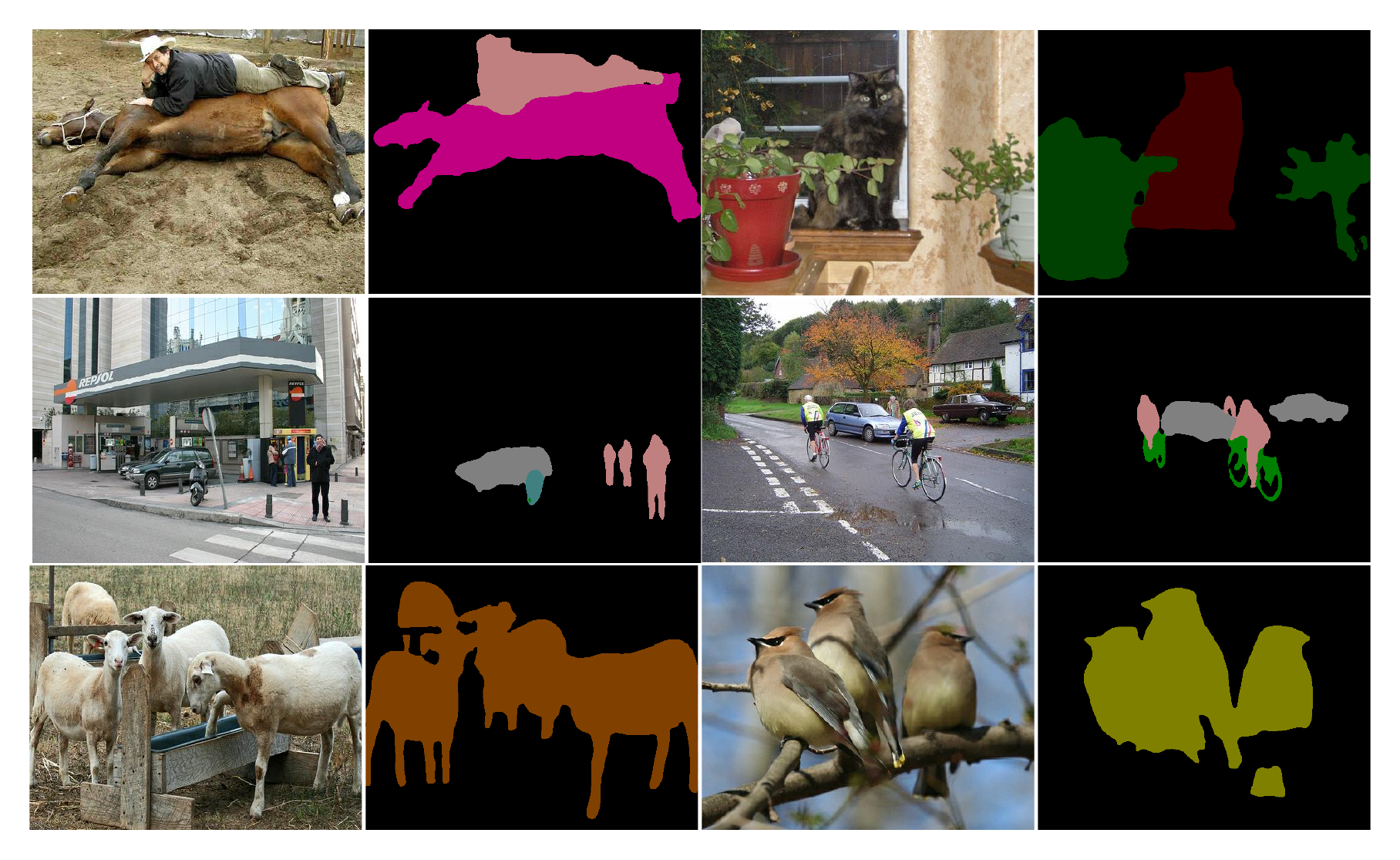}
\end{center}
\caption{Visualization results on Pascal VOC2012 test set.}
\label{VOC test set visual}
\end{figure}

%-------------------------------------------------------------------------
\subsection{Pascal VOC2012 Dataset}

\textbf{Pascal VOC2012:} The Pascal VOC2012 is well-known object segmentation dataset that includes 20 object categories and one background categories. There are 1,464 images for training, 1,449 images for evaluation and 1,456 images for testing. According to the common convention, we augment the training set by additional annotated VOC images provided in \cite{SBD} and the MS COCO dataset \cite{coco}.

We follow the same strategy to train our model on the Pascal VOC2012 dataset. The model used in Pascal VOC dataset is the same as the model used in cityscapes dataset. Table \ref{pascalvoctest} indicates the experiment results on Pascal VOC2012 testing set. Our proposed network achieves 88.4\% mIoU, outperforming other previous state-of-the-art methods. Compared with \cite{Deeplabv3+}, it increases 0.6\%. The results show that our method is better than the previous method for the variable scales object, such as aeroplane, bird, cow, horse, person.

\section{Conclusion}
In this paper, we have presented a Multi-Receptive Field Network for semantic segmentation. In order to capture multi-receptive field, the Multi-Receptive Field Module is proposed as a general module. Furthermore, MRFM-lite could achieve the trade-off between performance and speed. You can use different versions of the MRFM according to your needs. We have also provided an Edge Aware Loss which is effective in distinguishing the boundaries of object/stuff. At last, We hope our method could help all tasks of semantic segmentation improve the overall baseline.

{\small
\bibliographystyle{ieee_fullname}
\bibliography{egbib}
}

\end{document}